\documentclass[10pt,twocolumn,letterpaper]{article}

\usepackage{cvpr}              

\usepackage{graphicx}
\usepackage{amsmath}
\usepackage{amssymb}
\usepackage{booktabs}
\usepackage{colortbl}
\usepackage{multirow}
\usepackage{bbding}
\usepackage{makecell}
\usepackage{diagbox}
\usepackage{algorithm}
\usepackage{algorithmic}
\usepackage{fancyvrb}
\usepackage[dvipsnames]{xcolor}

\def\etal{\textit{et al}.}
\def\ie{\textit{i.e.}}
\def\eg{\textit{e.g.}}
\usepackage{xcolor}

\definecolor{citecolor}{RGB}{34,139,34}
\usepackage[pagebackref=true,breaklinks=true,letterpaper=true,colorlinks,citecolor=citecolor,bookmarks=false]{hyperref}
\PassOptionsToPackage{hyphens}{url}
\usepackage{hyperref}
\hypersetup{colorlinks=true}
\definecolor{mygray}{gray}{.9}
\newcommand{\app}{\raise.17ex\hbox{$\scriptstyle\sim$}}
\makeatletter\renewcommand\paragraph{\@startsection{paragraph}{4}{\z@}
  {.5em \@plus1ex \@minus.2ex}{-.5em}{\normalfont\normalsize\bfseries}}\makeatother

\usepackage[capitalize]{cleveref}
\crefname{section}{Sec.}{Secs.}
\Crefname{section}{Section}{Sections}
\Crefname{table}{Table}{Tables}
\crefname{table}{Tab.}{Tabs.}

\def\cvprPaperID{3172} 
\def\confName{CVPR}
\def\confYear{2022}

\begin{document}

\title{A Simple Data Mixing Prior for Improving Self-Supervised Learning}

\author{Sucheng Ren$^{1}$ ~~
Huiyu Wang$^{2}$ ~~
Zhengqi Gao$^{3}$ ~~
Shengfeng He$^{1*}$ ~~
Alan Yuille$^{2}$ \\
Yuyin Zhou$^{4}$ ~~
Cihang Xie$^{4}$\thanks{Corresponding authors: Shengfeng He (hesfe@scut.edu.cn), Cihang Xie (cixie@ucsc.edu)} \vspace{.3em}\\ 
$^1$ South China University of Technology \quad
$^2$Johns Hopkins University \\
$^3$ Massachusetts Institute of Technology \quad
$^4$UC Santa Cruz 
}
\maketitle

\begin{abstract}
Data mixing (\eg, Mixup, Cutmix, ResizeMix) is an essential component for advancing recognition models. In this paper, we focus on studying its effectiveness in the self-supervised setting. By noticing the mixed images that share the same source images are intrinsically related to each other, we hereby propose SDMP, short for \textbf{S}imple \textbf{D}ata \textbf{M}ixing \textbf{P}rior, to capture this straightforward yet essential prior, and position such mixed images as additional \textbf{positive pairs} to facilitate self-supervised representation learning. 

Our experiments verify that the proposed SDMP enables data mixing to help a set of self-supervised learning frameworks (\eg, MoCo) achieve better accuracy and out-of-distribution robustness. More notably, our SDMP is the first method that successfully leverages data mixing to improve (rather than hurt) the performance of Vision Transformers in the self-supervised setting. Code is publicly available at~\url{https://github.com/OliverRensu/SDMP}. 
\end{abstract}

\section{Introduction}
Data mixing can effectively improve recognition models. The very first data mixing strategy is introduced in \cite{mixup}, \ie, Mixup, which trains models on convex combinations of pairs of images and their labels.
This idea subsequently inspired several follow-ups, including mixing images and cropped patches \cite{cutmix}, mixing images and thumbnails \cite{xie2021cut}, and mixing among cropped patches \cite{takahashi2018ricap,bochkovskiy2020yolov4}.

\begin{figure}[t]
    \centering
      \includegraphics[width=.92\linewidth]{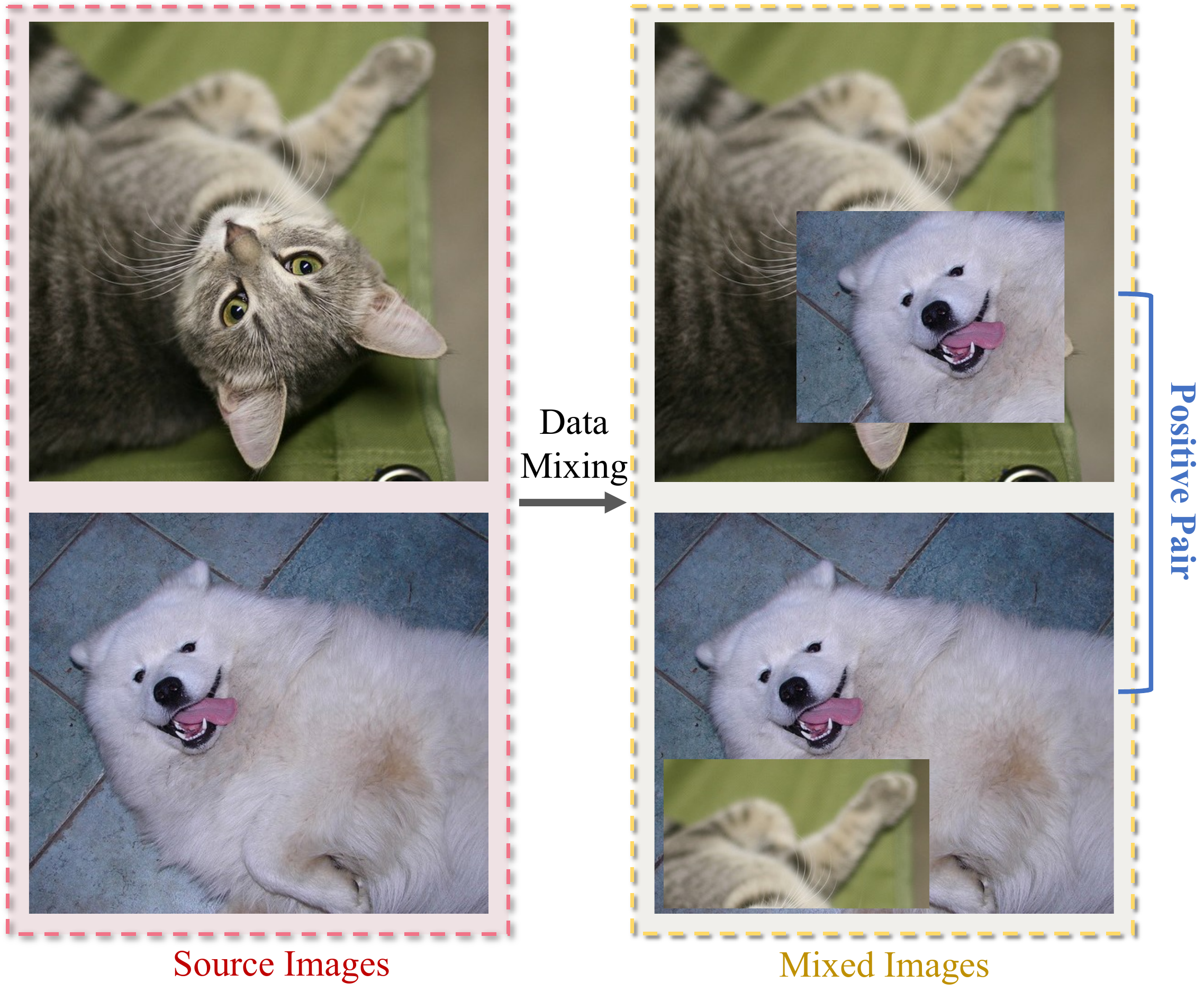}
      \vspace{-1em}
    \caption{For the mixed images that share the same source (\eg, a cat image and a dog image), they are semantically related and can be treated as additional positive pairs in self-supervised learning.}
    \vspace{-1em}
  \label{fig:cnn_inn_transf_bef_aft_distill}
\end{figure}

However, interestingly, data mixing plays little role in the recent surge of self-supervised learning. 
For instance, while na\"ively replacing original images with their mixed counterparts substantially improves Vision Transformers (ViTs) in the supervised setting \cite{touvron2020training}, it cannot improve ViTs under the self-supervised setting.
Though many efforts \cite{lee2021imix,verma2021towards,kalantidis2020hard} have been made recently by developing more sophisticated training strategies in data mixing, 
they are exclusively focusing on Convolutional Neural Networks (CNNs). As shown in Table~\ref{linear_results} in Section \ref{sec:exp},   these methods still fail to help (sometimes even hurt) ViTs \cite{dosovitskiy2020image}.

In this paper, \emph{we aim to develop a generic training strategy in data mixing that can improve the self-supervised representation learning of both CNNs and ViTs}.
By taking a closer look at the popular data mixing implementation where an image is mixed with another image \emph{that sampled from the same batch but in the flipped order}\footnote{The traditional data mixing implementation randomly samples two batches and then mixes them with each other; while this instantiation of data mixing only samples one batch and then mixes pairs of images from the same batch. It generally will not hurt performance, and is very popular in many libraries (\eg, timm \cite{rw2019timm}), due to its faster computation.}, we note such created mixed samples are inherently related in pairs (\eg, an example is provided in Figure \ref{fig:cnn_inn_transf_bef_aft_distill}). 
This indicates that now for one mixed image, there exist three related samples in the same training batch, \ie, a pair of source images and a mixed image created with a different mixing coefficient. This intrinsic relationship qualifies the pair of mixed images to be treated as additional positive samples in self-supervised learning to facilitate representation learning.

Motivated by the observation above, we hereby propose to leverage this \textbf{S}imple \textbf{D}ata \textbf{M}ixing \textbf{P}rior (dubbed \emph{SDMP}) to holistically model the relationship among samples for enhancing self-supervised learning. 
Different from previous methods \cite{lee2021imix,verma2021towards,kalantidis2020hard,li2020self,shen2020mix}, SDMP not only considers the relationships between source images and the mixed counterparts, but also encodes the connections between mixed samples in representation learning. We further enhance SDMP's representation learning by semantically weighting the loss to capture the relationships among samples accurately.

Our empirical results verify that the proposed SDMP successfully helps a set of self-supervised learning frameworks gain better accuracy on visual benchmarks and robustness on out-of-distribution samples, for both CNNs and ViTs. More essentially, we stress that our SDMP is the first strategy that enables data mixing to improve self-supervised ViTs. For example, by building upon the latest MoCo v3 \cite{mocov3}, while existing training strategies \cite{lee2021imix,verma2021towards,kalantidis2020hard} all hurt the top-1 ImageNet accuracy of ViT-S by 0.2\% - 1.6\%, SDMP successfully improves ViT-S by 0.6\%, attaining 73.8\% top-1 ImageNet accuracy. We hope our technical insights and empirical results will be helpful for future works on studying data mixing in self-supervised learning.

\section{Related Work}
\paragraph{Self-supervised learning.}
Self-supervised learning aims to let models acquire semantically meaningful representations without human annotations. Traditional pretext tasks include reconstruction by autoencoder~\cite{bengio2007greedy}, colorization~\cite{zhang2016colorful}, rotation prediction~\cite{gidaris2018unsupervised} or combinations of them~\cite{doersch2017multi,noroozi2018boosting}.

Contrastive learning, which aims to discriminate between different samples, is one of the most successful pretext tasks. Its core idea is to maximize the similarity of positive pairs and minimize the similarity of negative pairs.
However, discrimination based methods generally require a large amount of negative pairs, \eg, SimCLR~\cite{chen2020simple} needs a large training batch, MoCo~\cite{he2020momentum,chen2020improved} requires a memory bank, and others \cite{asano2019self,caron2018deep,zhan2020online} take a grouping/clustering.
Later works \cite{grill2020bootstrap,dino,chen2021exploring} successfully remove the need for negative samples, enabling small batch training. In this work, we focus on improving self-supervised learning by having extra positive pairs (generated in data mixing).

\paragraph{Data mixing.}
Mixup \cite{mixup} is the first work on data mixing, which convexly combines data pairs and their corresponding labels to regularize network training, inspiring numerous followups, including Cutout~\cite{Devries2017}, CutMix~\cite{cutmix}, SaliencyMix~\cite{uddin2020saliencymix} and PuzzleMix~\cite{kim2020puzzle}.

Recent works also introduce data mixing to regularize self-supervised learning. 
Verma \etal \cite{verma2021towards} utilize Mixup to create similar and dissimilar examples by mixing data samples differently, either at the input or hidden-state levels.
\cite{kim2020mixco,lee2021imix} explore the semi-contrastive encoding with a mixup of negative and positive pairs. 
Unlike previous works which exclusively focus on CNNs, we are the first to explore data mixing for improving ViTs under the self-supervised setting. We reveal properly modeling the relationships among mixed data (which was largely overlooked) can effectively strengthen self-supervised learning.

\paragraph{Transformers.}
Transformer \cite{Vaswani2017,devlin2018bert} is the de-facto standard for natural language processing tasks.
Recently, Dosovitskiy \etal \cite{dosovitskiy2020image} successfully introduced the pure Transformer architecture for computer vision, 
attaining competitive recognition performance compared to CNNs on a range of visual benchmarks.
Nonetheless, the original ViT training framework strongly demands hundreds of millions of images (\eg, the in-house JFT dataset \cite{sun2017revisiting}) in training. Touvron \etal \cite{touvron2020training} relax this learning constraint by incorporating a set of strong regularization techniques into ViT training framework, where data mixing (more specifically, Mixup and CutMix) plays a vital role. In this work, we are particularly interested in exploring the potential of data mixing in helping ViT in the self-supervised setting.

\section{Method}

\begin{figure*}[t]
    \centering
       \includegraphics[width=\textwidth]{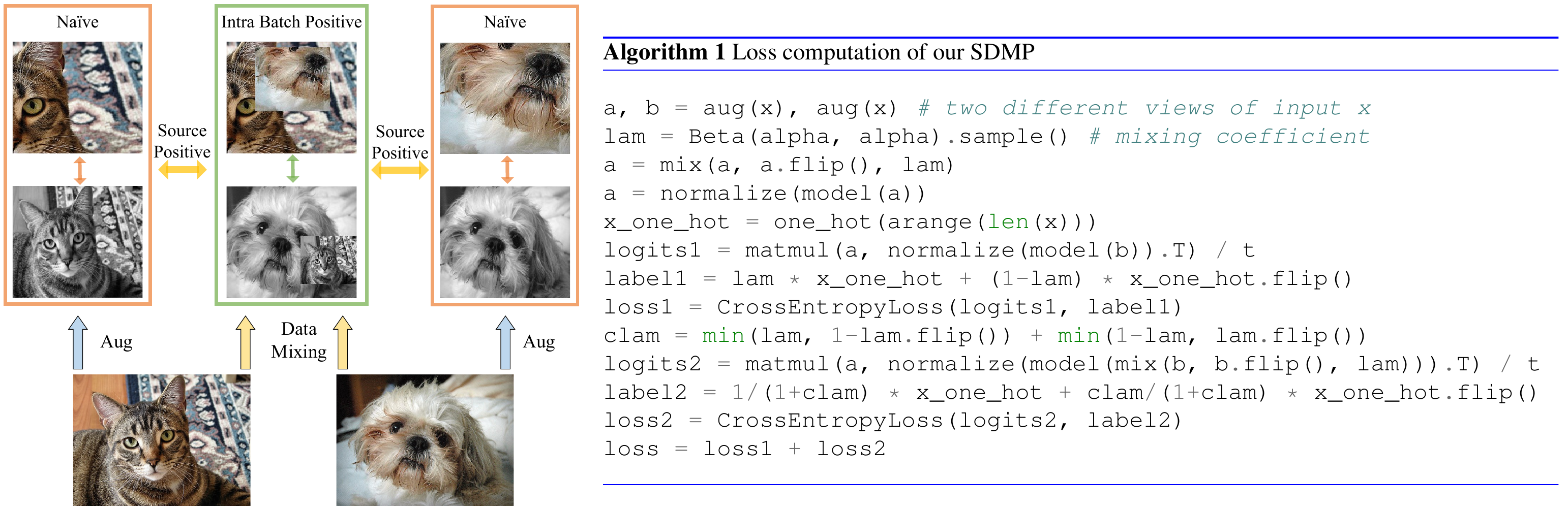}
    \vspace{-1.5em}
    \caption{Left panel: The positive pairs considered in self-supervised learning, \ie, the pair of different views (denoted as \emph{na\"ive}), the pair of the source image and the mixed image (denoted as \emph{source positive}), and the pair of the mixed images (denoted as \emph{intra batch positive}). 
    Right panel: the pseudo code of SDMP in PyTorch.}
  \label{fig:method}
\end{figure*}

\subsection{Which Images For Mixing?}
\label{sec:mix}
Traditionally, data mixing generates mixed data by mixing two randomly sampled images (usually drawn from two different min-batches). 
While in this work, we follow 
the mixing strategy applied in the widely used Deep Learning Package timm \cite{rw2019timm}---we mix the $i$-th image with another randomly selected $j$-th image that comes from the same batch. \emph{We by default set j equal to n-i where n is the number of images in this batch, for facilitating implementation}. We refer to this mixing strategy as \emph{intra-batch mixing}. Note this intra-batch mixing strategy naturally enables the mixed images to be related in pairs (as they share the same source images, see an example in Figure \ref{fig:cnn_inn_transf_bef_aft_distill}), which will facilitate our virtual label assignment introduced in Section \ref{sec:what_is_label}.

\subsection{How To Mix?}
\label{mixing}
To mix images, we mainly consider an element-wise data mixing method, Mixup~\cite{mixup}, and two regional data mixing methods, \ie, CutMix~\cite{cutmix} and ResizeMix~\cite{resizemix}.

\paragraph{Mixup.}
Mixup element-wisely combines two sample while preserving the original content of source data. Specifically, Mixup takes the weight $\lambda_i$ following a Beta distribution $\text{Beta}(\alpha,\alpha)$ and mixes the data as the following:
\begin{equation}
    \begin{split}
        x_i^{\text{mix}} = \lambda_i x_i +(1-\lambda_i)x_{n-i}, \\
    x_i^{\prime \text{mix}} = \lambda_i x^{\prime}_i +(1-\lambda_i)x^{\prime}_{n-i},
    \end{split}
\end{equation}
where $x_i$ and $x_{n-i}$ 
indicate the $i$-th and the $(n-i)$-th image; $x_i^{\prime}$ and $x_{n-i}^{\prime}$ are another view of the source images $x_i$ and $x_{n-i}$, created by data augmentation (\eg, color jittering).

\paragraph{CutMix.}
CutMix \cite{cutmix} combines data regionally by cropping and pasting a specific patch from one image to another, \ie,
the mixed data comes from one whole image and a local region of another image. Note that both Mixup and CutMix by default are included in ViT's training recipe \cite{touvron2020training} under the supervised training setting.

\paragraph{ResizeMix.}
One potential issue of CutMix is the cropped patch and the image itself could be label-irrelevant.
To let the cropped patch accurately deliver the corresponding semantics, ResizeMix \cite{resizemix} proposes to take the resized source image as the patch, and mix the data like the following:
\begin{equation}
    \begin{split}
    P_i=R(x_i, &H^{p}_i, W^{p}_i) \quad P_i^{\prime }=R(x^{\prime }_i, H^{p}_i, W^{p}_i),\\
    x_i^{\text{mix}} &= \text{Paste}(P_{n-i}, x_i), \\
    x_i^{\prime \text{mix}} &= \text{Paste}(P^{\prime }_{n-i}, x^{\prime }_i), \\
    \lambda_i &= 1 - \frac{W^p_{n-i}*H^p_{n-i}}{W_i * H_i},
    \end{split}
\end{equation}
where $R(\cdot, h, w)$ denotes the image resize function applied with the size of height $h$ and width $w$; $\text{Paste}(P, x)$ will paste the patch $P$ onto a random location of the image $x$; $H_i, W_i$ are the height and the width of the $i$-th image $x_i$; $H^{p}_i, W^{p}_i$ are the randomly sampled patch height and patch width for the $i$-th image $x_i$.

\subsection{What Is the Label?}
\label{sec:what_is_label}
Given true labels are not available in the self-supervised setting, we next show how to assign virtual labels accordingly.
Specifically, we showcase the virtual label assignments for two popular self-supervised learning frameworks.

\paragraph{Case 1: contrastive learning.}
\label{sec:moco}
Contrastive learning positions self-supervised learning as an instance classification task, assigning only one sample (created via data augmentation) as the positive pair and setting all the rest samples as negative. 
Our SDMP explicitly relaxes this assumption by introducing extra positive pairs.
Specifically,  we additionally assign the virtual positive labels to the following two groups: 1) the source data and the mixed counterparts; and 2) the pair of mixed data that share the same source data but with different mixing coefficients. This label assignment allows the model to learn to minimize the distance between more than just one pair.

We use the popular MoCo framework as the specific instantiation of contrastive learning. Firstly, to model the relationship between the source data and the mixed counterpart, the mix data $x_i^m$ will be fed into the encoder $f$ as the ``query'', and the other view of the source images, $x^{\prime }_i$ and $x^{\prime }_{n-i}$ (obtained via data augmentation), will be fed into the momentum encoder $f_k$ as the ``key''. The loss of this part (referred as source loss) can be written as:
\begin{equation}
\label{moco_source}
\begin{split}
    y_i^m=&f(x^m_i) \quad y^{\prime }_i=f_k(x^{\prime }_i) \quad y^{\prime }_{n-i}=f_k(x^{\prime }_{n-i})\\
    \mathcal{L}^{s}_{\text{MoCo}} =&-\lambda_i \log \frac{ \exp(<y_i^{m}, y^{\prime }_i>/\tau) }{\sum\limits_{j=0}^n \exp(<y_i^{m}, y^{\prime }_j>/\tau)} \\
    &- (1-\lambda_i)\log\frac{ \exp(<y_i^{m}, y^{\prime }_{n-i}>/\tau)}{\sum\limits_{j=0}^n \exp(<y_i^{m}, y^{\prime }_j>/\tau)}, 
\end{split}
\end{equation}
where $\tau$ is the temperature to normalize the output $y$; recall $\lambda_i$ is the mixing coefficient for creating the mix data $x^m_i$.

As mentioned in Section \ref{sec:mix}, we follow the intra-batch mixing strategy in timm \cite{rw2019timm} to mix one batch of data and its reversed-order counterpart accordingly. 
Note such mixed images are naturally related in pairs, \eg, $x_i^m$ and $x_{n-i}^{m}$ are related because both of them are created by mixing $x_i$ and $x_{n-i}$. In addition, by considering data augmentation, we propose to set $x_i^m$, $x_i^{\prime m}$ and $x_{n-i}^{\prime m}$ as the positive samples.
Therefore the loss here (referred to as mixing loss), which aims to learn from the relationship between mix data, could be written as:
\begin{equation}
\label{moco_mix}
    \begin{split}
        \lambda_{i}^{c} =& \min(\lambda_i, 1- \lambda_{n-i}) + \min(1-\lambda_i, \lambda_{n-i}), \\
    \mathcal{L}^{m}_{\text{MoCo}} = & - \frac{1}{1+\lambda_i^{c}} \log \frac{ \exp(<y_i^{m}, y^{\prime m}_i>/\tau)}{\sum\limits_{j=0}^n \exp(<y_i^{m}, y^{\prime m}_j>/\tau)} \\
    & - \frac{\lambda_i^{c}}{1+\lambda_i^{c}} \log \frac{\exp(<y_i^{m}, y^{\prime m}_{n-i}>/\tau)}{\sum\limits_{j=0}^n \exp(<y_i^{m}, y^{\prime m}_j>/\tau)},
    \end{split}
\end{equation}
where the coefficient $\lambda^c$ aims to capture the shared semantics between two mixed data. 

Note that Eq.~\eqref{moco_mix} can either be used as an extra view, or replace one of the symmetric view in MoCo. We hereby take the replace version to describe the total loss of our SDMP in MoCo:
\begin{equation}
    \mathcal{L} = \sum\limits_{i=0}^{n} \Big[\mathcal{L}_{\text{MoCo}}^s(x_i, x^{\prime }_i) + \mathcal{L}_{\text{MoCo}}^m(x^{\prime m}_i, x_i, x_{n-i}, x^m_i)\Big].
\end{equation}

\paragraph{Case 2:  knowledge  distillation.}
\label{sec:dino}
Recent works \cite{chen2021exploring,grill2020bootstrap,dino} show negative samples could be ``safely'' dropped in self-supervised learning. We hereby focus on studying the knowledge distillation framework introduced in DINO~\cite{dino}, where the student model (with the parameter $\theta_s$) is asked to match the output of the teacher model (with the parameter $\theta_t$). The corresponding distillation loss is:
\begin{equation}
\begin{split}
    H(P_t(x), P_s(x)) = - P_t(x) \log P_s(x),
\end{split}
\end{equation}
where $P_s(x)$ and $P_t(x)$ denote the output distribution (\ie, after a softmax function) of the student model and the teacher model, respectively.

In our data mixing case, there are two teachers that the student needs to distill from. The first one is distilling from the source teacher, which takes the source data as input:
\begin{equation}
\label{dino_source}
\begin{split}
    \mathcal{L}_{\text{DINO}}^{s} = H(\lambda_i P_t(x^{\prime }_i)+ (1-\lambda_i) P_t(x^{\prime }_{n-i}), P_s(x_i^{m})). 
\end{split}
\end{equation}

With Eq.~\eqref{dino_source}, we can minimize the distance between the mixed data and the two corresponding source data. Another teacher that needs to be distilled is the mix teacher, which take the mixed data as input:
\begin{equation}
\label{dino_mix}
    \begin{split}
     \mathcal{L}_{\text{DINO}}^{m} = \frac{1}{1+\lambda^c_i} H (P_t(x_i^{\prime m}),P_s(x_i^{m})) \\
    + \frac{\lambda^c_i}{1+\lambda^c_i} H (P_t(x_{n-i}^{\prime m}),P_s(x_i^{m})).
    \end{split}
\end{equation}

With Eq.~\eqref{dino_mix}, the student can learn the output distribution of the mixed data from the mix teacher. 
Note that both the source teacher and the mix teacher share the same network parameter $\theta_t$, which is updated by using the exponential moving average of the student parameter $\theta_s$.

To sum up, the total loss of our SDMP in DINO is:
\begin{equation}
    \mathcal{L} = \mathcal{L}_{\text{DINO}} + \mathcal{L}_{\text{DINO}}^m + \mathcal{L}_{\text{DINO}}^s,
\end{equation}
where $\mathcal{L}_{\text{DINO}}$ is the original DINO loss that applied to the teacher-student pairs without data mixing.
\section{Experiment}
\label{sec:exp}
This section evaluates the effectiveness of SDMP on a set of standard visual benchmarks, including ImageNet \cite{deng2009imagenet} and CIFAR-10/100 \cite{Krizhevsky09learningmultiple}.
Given no prior works successfully enable data mixing to improve self-supervised ViTs, 
we take ViT-S as the major backbone in experiments. 
The input patch size of ViT-S is $16\times 16$, therefore resulting in a sequence length of 196 for a $224\times 224$ input image. ViT-S has 12 transformer blocks, and the feature dimension of each block is 384. For data augmentation, we follow the settings in BYOL~\cite{grill2020bootstrap}, which includes random resize and crop, color jittering, Gaussian Blurring, and solarization. We use Adam with weight decay as the optimizer \cite{loshchilov2017decoupled}, and set the learning rate $lr$ following the linear scaling rule \cite{goyal2017accurate}: $lr = 0.0005*\text{batchsize}/256$; our default training batch size is 1024.

We set contrastive learning and knowledge distillation as the default pretext tasks in self-supervised learning, and comprehensively measure the quality of learned representations via linear evaluation, end-to-end finetuning, and semi-supervised learning. 
In addition, we report model robustness on multiple out-of-distribution benchmarks, including ImageNet-A~\cite{imageneta}, ImageNet-R~\cite{imagenetr} and ImageNet-C~\cite{imagenetc}.

\subsection{Classification on ImageNet}

\subsubsection{Linear Evaluation}
\label{sec:linear_finetune}
\begin{table}
\centering
\resizebox{\linewidth}{!}{
\begin{tabular}{l|c|c|c|c}
\toprule
Method      & Model &Param. &Epoch & Top-1 (\%) \\
\midrule
\midrule
Supervised~\cite{touvron2020training}  & ViT-S&21M &300   & 79.8   \\
BYOL~\cite{grill2020bootstrap}       & ViT-S   &21M &300  & 71.4  \\
MoCo v2~\cite{chen2020improved}     &  ViT-S  &21M &300    & 72.7   \\
SwAV~\cite{swav}        & ViT-S  &21M &300     & 73.5   \\
MoCo v3~\cite{mocov3}      &  ViT-S  &21M &300    &    73.2    \\
+ imix*~\cite{lee2021imix}      &  ViT-S  &21M &300    &    71.6    \\
+ DACL*~\cite{verma2021towards}      &  ViT-S  &21M &300    &    72.3    \\
+ MoChi*~\cite{kalantidis2020hard}      &  ViT-S  &21M &300    &   73.0    \\
\rowcolor{mygray}
+ SDMP (ours) & ViT-S   &21M &300   &    73.8   \\
DINO~\cite{dino}        & ViT-S   &21M &300    & 76.0   \\
\rowcolor{mygray}
+ SDMP (ours)  & ViT-S   &21M &300  & 76.4  \\
\midrule
MoCo v3 & ViT-B   &85M &300   &    76.7  \\
\rowcolor{mygray}
+ SDMP (ours) & ViT-B   &85M &300   &    77.2   \\
\midrule
SimCLR~\cite{chen2020simple} &  Res50  &23M &200  & 60.6   \\
BYOL~\cite{grill2020bootstrap} &   Res50 &23M &200  & 61.9   \\
SwAV~\cite{swav} &   Res50 &23M &800  & 75.3  \\
MoCo v1~\cite{he2020momentum} &   Res50  &23M &200  & 60.6   \\
MoCo v2~\cite{chen2020improved} &   Res50  &23M &800  & 71.1   \\
MoCo v3~\cite{mocov3}      &  Res50  &23M &300    &    72.8    \\
 + i-mix*~\cite{lee2021imix}      &  Res50  &23M &300    &    72.8    \\
\rowcolor{mygray}
 + SDMP (ours) & Res50   &23M &300   &    73.5    \\
\bottomrule
\end{tabular}}
\vspace{-.7em}
\caption{The ImageNet performance of different pre-training methods under the linear evaluation protocol. Our SDMP consistently brings improvements over the baselines for both ViTs and CNNs. * indicate our reproduced results.}
\vspace{-1em}
\label{linear_results}
\end{table}

Linear evaluation \cite{moco,dino} accesses the performance of self-supervised learning by freezing all parameters in the backbone network and only training a linear classifier on top of it. Following the setups in the prior work~\cite{moco}, we only take resize, crop and random flipping to augment training images. Due to the variance of feature space in different self-supervised methods, we will adjust the initial learning rate accordingly. Note that most self-supervised ViTs only take the last class token for linear evaluation, while DINO takes the last four class tokens; we follow this setup in our DINO experiments. Moreover, when applying the proposed SDMP to MoCo, we randomly replace one of the symmetric views with the mixed data; when applying the proposed SDMP to DINO, we randomly replace half local views with the mixed data. This replacing strategy ensures SDMP will not bring extra computations to train student models.

\paragraph{Main results.} We report the ImageNet linear evaluation results in Table \ref{linear_results}. Firstly, we note that the proposed SDMP can bring consistent improvements over MoCo and DINO. For examples, compared to the latest MoCo v3 baseline, SDMP can further boost the top-1 accuracy of ViT-S by 0.6\% (\ie, from 73.2\% to 73.8\%); for DINO, SDMP increases the top-1 accuracy of ViT-S to 76.4\% (+0.4\%, from 76.0\%). In contrast, for existing training strategies that use data mixing, including i-mix \cite{lee2021imix}, DACL \cite{verma2021towards} and MoChi \cite{kalantidis2020hard}, we note they all fail to help ViT gain stronger performance over the vanilla MoCo v3 baseline, decreasing the top-1 accuracy by 0.2\% - 1.6\%. Moreover, we verify that SDMP scales well with large ViTs, \ie, it successfully helps ViT-B beat the vanilla MoCo v3 baseline by 0.5\% (\ie, from 76.7\% to 77.2\%).

Lastly, we observe that SDMP also helps CNNs gain better performance in self-supervised learning. For example, with ResNet-50, SDMP effectively improves the MoCo v3 baseline by 0.7\% (from 72.8\% to 73.5\%), while the existing approaches like i-mix hardly bring in extra accuracy gain.

\subsubsection{End-to-End Fintuning}
\label{sec:e2e-finetune}

\begin{table}
\centering
\resizebox{0.82\linewidth}{!}{
\begin{tabular}{l|c|c|c|c}
\toprule
Method      & Model &Param. &Epoch & Top-1 \\
\midrule
\midrule
Supervised  & ViT-S&21M &100   & 75.8   \\
Supervised  & ViT-S&21M &300   & 79.8   \\
\midrule
MoCo v3      &  ViT-S  &21M &100    &    78.7    \\
\rowcolor{mygray}
+ SDMP & ViT-S   &21M &100   &    79.1    \\
\midrule
DINO        & ViT-S   &21M &100    & 79.7   \\
\rowcolor{mygray}
+ SDMP   & ViT-S   &21M &100  & 80.0  \\
\bottomrule
\end{tabular}
}
\vspace{-.7em}
\caption{End-to-end fintuning on ImageNet. For self-supervised methods, all models here are pre-trained for 300 epochs. The ``Epoch'' in the table refers to the number of fintuning epochs.}
\vspace{-.6em}
\label{E2E}
\end{table}

\begin{table}[htb!]
\centering
\resizebox{0.82\linewidth}{!}{
\begin{tabular}{l|c|c|c|c}
\toprule
Method      & Model &Param. & ~10\%~ & ~1\%~ \\
\midrule
\midrule
MoCo v3      &  ViT-S  &21M &66.7    &    54.4    \\
\rowcolor{mygray}
+ SDMP & ViT-S   &21M &67.4   &   55.5   \\
\midrule
DINO        & ViT-S   &21M &67.2    &  55.6  \\
\rowcolor{mygray}
+ SDMP   & ViT-S   &21M &68.0  &  56.3 \\
\bottomrule
\end{tabular}
}
\vspace{-.7em}
\caption{Semi-supervised learning on ImageNet with 10\% and 1\% labeled data. All methods are pre-trained for 300 epochs.}
\vspace{-.7em}
\label{semi}
\end{table}

We next follow the training recipe in DeiT~\cite{touvron2020training} to finetune pre-trained ViTs. Specifically, these ViT are first pre-trained for 300 epochs in the self-supervised setting, and then finetuned for 100 epochs in the supervised setting. We additionally consider the pure supervised training as a strong baseline for comparison.

We present the comparisons in Table \ref{E2E}. Firstly, compared to the vanilla self-supervised training baselines,  SDMP brings consistent improvements. For example, with ViT-S, it beats the MoCo v3 baseline by 0.4\% and the DINO baseline by 0.3\%. More interestingly, when comparing to the strong supervised training baselines, we note SDMP a) substantially outperforms the 100-epoch supervised training baseline; and b) can closely match, or even outperforms, the strong 300-epoch supervised training baseline.

\subsubsection{Semi-Supervised Learning}
We now follow the semi-supervised learning protocol in \cite{chen2020big}, where the pre-trained will be finetuned with only a small portion of ImageNet, \eg, 1\% data or 10\% data. 
The results are reported in Table \ref{semi}. Firstly, our SDMP can consistently outperform the MoCo v3 baseline and the DINO baseline in both the 1\% data setting and the 10\% data setting.
We note the performance gap between SDMP and the baseline tends to become larger if less data is used for finetuning. 
For example, with MoCo v3, the performance gap is 0.4\% with 100\% data (the end-to-end finetuning setting in Section \ref{sec:e2e-finetune}), 0.7\% with 10\% data, and 1.1\% with 1\% data. This result suggests that SDMP can help self-supervised learning generalize better in the small labeled-data regime.

\subsubsection{Robustness on Out-of-Distribution Datasets}
\begin{table}
\centering
\resizebox{0.82\linewidth}{!}{
\begin{tabular}{l|c|c|c|c}
\toprule
\multirow{2}{*}{Method}    & ImageNet  & A   & R  & C  \\
&(\%)&(\%)&(\%)&(\%) \\
\midrule
\midrule
MoCo v3   &       78.7   &  18.1 & 42.1  &  52.9 \\ 
\rowcolor{mygray}
+ SDMP       &    79.1      & 18.9  & 42.8  & 53.4  \\ 
\midrule
DINO        &    79.7     &  20.0 &  44.9 &  54.7 \\
\rowcolor{mygray}
+ SDMP       &      80.0    & 21.1   & 45.3 & 55.0  \\ 
\bottomrule
\end{tabular}
}
\vspace{-.7em}
\caption{Performance on ImageNet and out-of-distribution datasets. ``A'', ``R'', ``C'' refer to ImageNet-A \cite{imageneta}, ImageNet-R \cite{imagenetr}, and ImageNet-C \cite{imagenetc}, respectively. Note that when measuring performance on ImageNet-C, we directly use top-1 accuracy (the higher the better) as the evaluation metric, rather than using ``mCE'' (the lower the better) as originally defined in \cite{imagenetc}.}
\vspace{-.8em}
\label{OOD}
\end{table}

We hereby evaluate model robustness on out-of-distribution data. Specifically, we test the performance on perturbed versions of ImageNet, \ie, natural adversarial examples (ImageNet-A~\cite{imageneta}), semantic shifts (ImageNet-R~\cite{imagenetr}), and common image corruption (ImageNet-C~\cite{imagenetc}). 

As shown in Table \ref{OOD}, our SDMP consistently improves the performance of the MoCo v3 baseline and the DINO baseline on these out-of-distribution tests.
Notably, this robustness improvement could be much more substantial than the performance gain in the standard ImageNet benchmark.
For instance, with DINO, our SDMP ``merely'' improve the ImageNet end-to-end finetuning result by 0.3\%, whereas its gain on ImageNet-A is 1.1\%.
These results suggest that SDMP can be an ideal addition to existing self-supervised learning frameworks for enhancing model robustness.

\subsection{CIFAR-10/100 Results}
\label{sec:cifar}
In addition to ImageNet, we study the popular CIFAR-10 and CIFAR-100 datasets~\cite{Krizhevsky09learningmultiple} to verify the generalization of SDMP. 
Given it is extremely non-trivial to directly pre-train ViTs on the small CIFAR-10/100 dataset, we hereby exclusively focus on studying CNN architectures, more specifically, the ResNet family~\cite{he2016deep}. We will study ViTs + SDMP on the other small dataset in Section \ref{sec:vit_on_small_dataset}.

\paragraph{Experiment setup.} CIFAR-10/100 contain 32$\times$32 small size images with 10/100 classes, respectively. Both datasets contain 50000 training images and 10000 testing images. We select ResNet-50 and ResNet-101 as the main CNN architectures for pre-training. We slightly modify the ResNet architecture to make it more suitable for the CIFAR-10/100 training:
for the first convolution layer, we change the kernel size from $7\times 7$ to $3\times 3$ and reduce the stride from 2 to 1; we also remove the max pooling layer that was originally placed right after the first convolution layer. We select the latest MoCo v3 as our self-supervised learning framework.  
In addition, we re-implement i-mix in MoCo v3 as a strong baseline for comparison.

\paragraph{CIFAR-10.} The linear evaluation results on CIFAR-10 are shown in the fourth column of Table \ref{CIFAR}. Firstly, compared to the vanilla MoCo v3 baseline, we note applying data mixing always yields better CIFAR-10 performance. Secondly, though both approaches enhance the MoCo v3 baseline, we note that SDMP consistently attains higher performance than i-mix. For example, by pre-training ResNet-50 for 200 epochs, i-mix improves the vanilla baseline by 2.1\%  while SDMP yields a large performance improvement of 3.0\%. 

\begin{table}[t!]
\centering
\begin{tabular}{l|c|c|c|c}
\toprule
Method      & Model &Epoch &CIFAR10 & CIFAR100 \\
\midrule
\midrule
MoCo v3      &  Res50  &200 &86.5    &    63.2   \\
+ i-mix      & Res50  &200 &88.6    &    66.1    \\
\rowcolor{mygray}
+ SDMP      &  Res50  &200 &89.5    &    68.2    \\
\midrule
MoCo v3      &  Res50  &2000 &93.7    &    69.0   \\
+ i-mix      & Res50  &2000 &95.4    &    77.3    \\
\rowcolor{mygray}
+ SDMP      &  Res50  &2000 &95.8    &    78.7    \\
\midrule
MoCo v3      &  Res101  &200 &86.4    &    63.3   \\
+ i-mix      & Res101  &200 &89.4    &    67.7    \\
\rowcolor{mygray}
+ SDMP      &  Res101  &200 &90.0    &    69.7    \\
\midrule
MoCo v3      &  Res101  &2000 &93.8    &    68.5   \\
+ i-mix      & Res101  &2000 &95.8    &    78.4    \\
\rowcolor{mygray}
+ SDMP      &  Res101  &2000 &95.8    &    80.0    \\
\bottomrule
\end{tabular}
\vspace{-.7em}
\caption{The CIFAR-10/100 performance of different pre-training methods under the linear evaluation protocol. We note SDMP consistently yields the best performance among all settings.}
\vspace{-1em}
\label{CIFAR}
\end{table}

\paragraph{CIFAR-100.} 
We report the linear evaluation results on CIFAR-100 in the last column of Table \ref{CIFAR}. Similar to the results in CIFAR-10, applying data mixing also helps MoCo v3 in CIFAR-100. We note the performance improvement is particularly substantial when you train with a larger model for longer epochs, \eg, by pre-training ResNet-101 for 2000 epochs, both i-mix and SDMP can outperform the vanilla MoCo v3 by at least \app 10.0\%. We conjecture this is because CIFAR-100 is a relatively smaller dataset (\eg, only 500 training samples per class); therefore, data augmentation is strongly demanded in training to help larger models generalize better. Next, we note SDMP consistently yields better performance than i-mix. For example, for both ResNet-50 and ResNet-101, SDMP beats i-mix by \app 2\% in the 200-epoch training, and by \app 1.5\% in the 2000-epoch training.

The results above suggest that the proposed SDMP can effectively generalize to CIFAR-10 and CIFAR-100. Moreover, we note the proposed SDMP can consistently outperforms i-mix \cite{lee2021imix} in all settings, corroborating our observation on ImageNet (in Section \ref{sec:linear_finetune}) that encoding the relationship between mixed data is essential for enhancing self-supervised learning.

\subsection{Ablation Study}
\subsubsection{On the importance of $\lambda$}

\begin{table}
\centering
\resizebox{0.92\linewidth}{!}{
\begin{tabular}{l|c|c|c|c}
\toprule
Method      & $\lambda_i$ &$\lambda_i^c$&Linear & Finetuning \\
\midrule
\midrule
MoCo v3     & None &None &73.2    &    78.7   \\
\midrule

+ SDMP & Static  &Rand. &71.5  &    78.0 \\
+ SDMP & Rand.   &Static &72.5   &    78.4  \\
\rowcolor{mygray}
+ SDMP & Rand.   &Rand. &73.8   &   79.1  \\
\midrule
DINO        & None   &None &76.0    & 79.7 \\
\midrule
+ SDMP   & Static   &Rand. &75.5  & 79.4  \\
+ SDMP   & Rand.   &Static &76.1  & 79.9 \\
\rowcolor{mygray}
+ SDMP   & Rand.   &Rand. &76.4  & 80.0  \\
\bottomrule
\end{tabular}
}
\vspace{-.7em}
\caption{The ImageNet performance of different setups of $\lambda$. ``None'' refers to no data mixing is applied. ``Rand.'' is the default setup in SDMP, which calculate $\lambda_i$ or $\lambda_i^c$ based on the randomly sampled $\lambda$.  ``Static'' refers to the setting that $\lambda_i$ or $\lambda_i^c$ in the loss is a pre-defined constant and irrelevant to $\lambda$.}
\vspace{-0.8em}
\label{weight}
\end{table}

We sample $\lambda$ from Beta distributions as the coefficient when mixing two images. We treat this $\lambda$ as the prior in the loss during pre-training. In this part, we ablate how different setups of $\lambda$ affect model performance.

\paragraph{Static weight.} There are two parts in the training loss related to the data mixing coefficient $\lambda$: the source loss (\eg, Eq. \eqref{moco_source} and Eq. \eqref{dino_source}) and the mixing loss (\eg, Eq. \eqref{moco_mix} and Eq. \eqref{dino_mix}). To check whether $\lambda$ is a useful prior, we manually opt out the prior when computing the loss. Specifically, we keep the randomly sampled $\lambda$ in data mixing;
whereas for the loss computation, we could 1) set $\lambda_i = 0.5$ for ablating the source loss (\eg, now both $\lambda_i$ and $1-\lambda_i$ equal to 0.5), or 2) set $\lambda^c_i=1$ for ablating the mixing loss (\eg, now both $\frac{1}{1+\lambda^c_i}$ and $\frac{\lambda^c_i}{1+\lambda^c_i}$ equal to 0.5).
We referred to this weight assigning strategy as ``static weight''; for the original weight assigning strategy introduced in Section \ref{sec:what_is_label}, we refer to it as ``random weight''.

We report the ImageNet evaluation results in Table \ref{weight}. Firstly, when applying ``static weight'' in the source loss (\ie, $\lambda_i = 0.5$), we note the performance of both DINO and MoCo v3 substantially drops in both the linear evaluation and the end-to-end finetuning. Moreover, we note the resulted models even perform much worse than the vanilla MoCo v3 or DINO baselines. But meanwhile, one interesting observation is that, compared to MoCo v3, DINO can more robustly cope with ``static weight''. For example, by changing from ``random weight'' to ``static weight'', the linear evaluation accuracy largely drops by 2.3\% in MoCo v3 (from 73.8\% to 71.5\%), while such drop in DINO is only 0.9\% (from 76.4\% to 75.5\%).

Next, we ablate the effect of ``static weight'' in the mixing loss (\ie, $\lambda_i^c = 1$). For MoCo v3, applying ``static weight'' always hurts performance. While for DINO, we find that ``static weight'' and ``random weight'' lead to similar accuracy, \eg, 79.9\% \vs 80.0\% when finetuning the whole model. We conjecture this phenomenon should be attributed to the intra-batch similarity, as the intra-batch sample pair we built shares the same source data but only with different mixing coefficients, therefore by even using a constant $\lambda^c_i$ can somewhat capture the overlapped semantics.

\paragraph{Per-sample weight vs. per-batch weight.} By default, we assign each sample with a randomly sampled mixing coefficient $\lambda$. An alternative strategy is to assign a shared mixing coefficient for the entire training batch, namely, setting $\lambda_1=\lambda_2= ... =\lambda_n$ for the training batch with $n$ samples. The results are reported in Table \ref{shared_weight}. We note that taking a shared mixing coefficient leads to a performance decrease, \eg, the accuracy drops by 1.5\% in the linear evaluation and 0.9\% in end-to-end finetuning for MoCo v3.
We conjecture this is because the same mixing pattern is shared across the whole training batch, therefore weakening the training regularization brought by data mixing.
These results suggest that assigning the per-sample $\lambda$ is more effective than assigning the per-batch $\lambda$ for mixing the data.

\begin{table}
\centering
\begin{tabular}{l|c|c|c|c}
\toprule
Method      & $\lambda_i$ &$\lambda_i^c$&Linear & Finetuning \\
\midrule
\midrule
MoCo v3     & None &None &73.2    &    78.7   \\
+ SDMP & Shared & Shared &72.3   &   78.2 \\
\rowcolor{mygray}
+ SDMP & Ind. & Ind. &73.8   &   79.1  \\
\midrule
DINO        & None   &None &76.0    & 79.7 \\
+ SDMP   & Shared & Shared &74.5  &79.0  \\
\rowcolor{mygray}
+ SDMP   & Ind. & Ind. &76.4  & 80.0  \\
\bottomrule
\end{tabular}
\vspace{-.7em}
\caption{The ImageNet performance of using independent (per-sample) $\lambda$ (denoted as ``Ind.'') or shared (per-batch) $\lambda$ across the whole training batch (denoted as ``Shared'').}
\vspace{-1em}
\label{shared_weight}
\end{table}

\subsubsection{Data Mixing Strategies}
For each training batch, the default setup in SDMP is to select a data mixing strategy from the set \{Mixup, Cutmix and ResizeMix\} uniformly at random. To ablate the effects of applying different data mixing strategies, we then compare our default setup with two additional settings: 1) applying exclusively with the element-wise data mixing Mixup; and 2) applying exclusively with the regional data mixing Resizemix. We report the results in Table \ref{mix}. We note that: 1) our SDMP consistently outperforms the vanilla MoCo v3 or DINO baselines, even if only one data mixing strategy is applied; 2) our SDMP achieves the best results when \{Mixup, Cutmix and ResizeMix\} are all used. 
\begin{table}[h!]
\centering
\vspace{-.55em}
\resizebox{\linewidth}{!}{
\begin{tabular}{l|c|c|c|c}
\toprule
Method      & Model &Mixing &Epoch & Top-1 (\%) \\
\midrule
\midrule
MoCo v3    &  ViT-S  &None &100    &   64.7   \\
+ SDMP    &  ViT-S  &Mixup &100    &   65.1   \\
+ SDMP & ViT-S   &Resizemix &100   &    65.4   \\
\rowcolor{mygray}
+ SDMP & ViT-S   &All &100   &  65.5     \\
\midrule
\midrule
DINO    &  ViT-S  &None &100    &   73.8   \\
+ SDMP    &  ViT-S  &Mixup &100    &   74.3    \\
+ SDMP & ViT-S   &Resizemix &100   &   74.4   \\
\rowcolor{mygray}
+ SDMP & ViT-S   &All &100   &  74.4    \\
\bottomrule
\end{tabular}}
\vspace{-.8em}
\caption{Ablations of different data mixing strategies. }
\label{mix}
\vspace{-1em}
\end{table}

\subsubsection{Extra View Version \vs Replace Version in MoCo}
The MoCo v3 framework default sees two differently augmented views of the same input. To additionally incorporate data mixing, the mixed data can either be used to replace one of the existing views or form an extra view. 
To ensure a fair comparison among different strategies, we set \emph{the total training epoch to be the same as the vanilla  MoCo v3 baseline for the replace version, but reduce the total training epoch by 1/3 for the extra view version}. This is because, compared to the vanilla MoCo v3 baseline and the replace version, the extra view version requires three views (instead of two views) from the same training sample. 
Table~\ref{extra_replace} reports the corresponding ImageNet performance. We can observe that both the replace version and the extra view version outperform the vanilla MoCo v3 baseline.

\begin{table}[h!]
\vspace{-.5em}
\centering
\resizebox{0.92\linewidth}{!}{
\begin{tabular}{l|c|c|c}
\toprule
Method      & Model  &Epoch & Top-1 (\%) \\
\midrule
\midrule
MoCo v3     &  ViT-S   &150    &    66.7    \\
+ SDMP (Replace) & ViT-S    &150   &  67.4   \\
+ SDMP (Extra)    &  ViT-S  &100    &   67.5    \\
\bottomrule
\end{tabular}
}
\vspace{-.8em}
\caption{Comparisons among the extra view version, the replace version and the vanilla MoCo v3 baseline on ImageNet.}
\label{extra_replace}
\vspace{-1.55em}
\end{table}

\subsubsection{Local Views in DINO}

\begin{table}[b!]
\vspace{-1.2em}
\centering
\resizebox{0.96\linewidth}{!}{
\begin{tabular}{l|c|c|c|c|c}
\toprule
\multirow{2}*{Method}      & Global &Global&Local & Local & Top-1\\
     & Clean &Mixed &Clean & Mixed & (\%)\\
\midrule
DINO       & 2   & \XSolidBrush &  \XSolidBrush  & \XSolidBrush&67.8 \\
+ SDMP        & 1   &1 &\XSolidBrush  & \XSolidBrush &64.1\\
\midrule
DINO       & 2   & \XSolidBrush &  2  & \XSolidBrush& 71.5\\
+ SDMP        & 2   &\XSolidBrush  &  1& 1 &70.9\\ 
\midrule
DINO       & 2   & \XSolidBrush &  6  & \XSolidBrush &73.8\\
+ SDMP        & 2   &\XSolidBrush  &  3& 3 &74.0\\ 
\midrule
DINO       & 2   & \XSolidBrush &  8  & \XSolidBrush &74.0\\
+ SDMP        & 2   &\XSolidBrush  &  4& 4 &74.4\\ 
\bottomrule
\end{tabular}
}
\vspace{-.7em}
\caption{Ablating the effects of local views on ImageNet accuracy. All models are pre-trained for 100 epochs. Global/Local Clean denotes no data mixing is applied to global/local views. Global/Local Mixed denotes global/local views with data mixing.}
\label{local_crop}
\end{table}

One of the most important contributions in DINO is encouraging ``local-to-global'' correspondences \cite{dino}, \ie, in addition to two global views (at resolution $224^2$ covering a large image region), DINO additionally introduces several local views (at resolution $96^2$ covering only a small image region) in training. 
We hereby investigate the importance of the number of local views on SDMP. Specifically, when applying SDMP to DINO, we always replace half of the local views with mixed data; for the extreme case that no local views exist, we replace one of the global views with our mixed data. The ImageNet linear evaluation results are reported in Table \ref{local_crop}. We note SDMP cannot work well, and sometimes even underperforms the vanilla DINO baseline, when no local views or only a few local views exist. Interestingly, by increasing the number of local views, SDMP begins to bridge such a performance gap, eventually outperforming the vanilla DINO baseline. For example, by training with eight local views, SDMP beats the vanilla DINO baseline by 0.4\% (74.4\% \vs 74.0\%). These results suggest that having enough local views are essential for ensuring the improvements brought by data mixing.

\subsubsection{Generalization on Small Datasets with ViTs}
\label{sec:vit_on_small_dataset}
Section \ref{sec:cifar} demonstrates ResNet + SDMP can generalize well to small-scale datasets. We hereby verify if this conclusion holds for ViT + SDMP. We choose the relatively small dataset, \ie, ImageNet-100, for this purpose. As shown in Table \ref{Image100}, SDMP consistently improves MoCo v3 (from 79.1\% to 81.8\%) and DINO (from 82.0\% to 83.2\%), showing its effectiveness at different data scale regime with ViTs.

\begin{table}[h]
\centering
\vspace{-.5em}
\resizebox{0.87\linewidth}{!}{
\begin{tabular}{l|c|c|c|c}
\toprule
Method      & Model &Param. &Epoch & Linear \\
\midrule
\midrule
Supervised      &  ViT-S  &21M &300    &    88.0   \\
\midrule
MoCo v3      &  ViT-S  &21M &300    &    79.1   \\
\rowcolor{mygray}
+ SDMP & ViT-S   &21M &300   &  81.8  \\
\midrule
DINO        & ViT-S   &21M &300    & 82.0  \\
\rowcolor{mygray}
+ SDMP   & ViT-S   &21M &300  & 83.2 \\
\bottomrule
\end{tabular}
}
\vspace{-.7em}
\caption{Linear evaluation on the small-scale ImageNet-100.}
\vspace{-1em}
\label{Image100}
\end{table}

\section{Conclusion}
In this paper, we develop a generic training strategy for enabling data mixing to effectively help self-supervised training, especially with Vision Transformers. By following the intra-batch data mixing strategy in timm \cite{rw2019timm}, we propose SDMP to capture the intrinsic relationships between mixed data in a precise manner. 
Experiments show that our method brings consistent improvements, and is compatible with various self-supervised learning frameworks, architectures, and datasets.

\paragraph{Discussion \& Limitation}
This work mines the intra-batch relationship between mixed samples to help self-supervised learning.
Future work could examine how to integrate our method into the recent self-supervised masked image modeling methods \cite{bao2021beit,he2021masked,zhou2021ibot}. In addition, due to computational limitations, our experiments are mainly built upon the small-sized ViT (\ie, ViT-S); future works could verify the effectiveness of our method at a larger scale.

\vspace{.37em}
{\small
{\noindent {\bf Acknowledgement}: This work is supported by a gift from Open Philanthropy, ONR N00014-21-1-2812, and Google Cloud Research Credits program.}
}
{\small
\bibliographystyle{ieee_fullname}
\bibliography{egbib}
}

\end{document}